\documentclass[conference,a4paper]{IEEEtran}
\usepackage{cite}
\usepackage{graphicx}
\graphicspath{{./images/}}
\usepackage[cmex10]{amsmath}
\usepackage{siunitx}
\usepackage{algorithmic}
\usepackage{booktabs}
\usepackage{siunitx}
\usepackage{float}
\usepackage{pgfplots}
\pgfplotsset{width=3.5in,compat=1.15}
\usepackage{tikz}
\usepackage{tikz-3dplot}
\tikzstyle{every pin}=[font=\footnotesize,align=center]
\usepackage{tabularx,booktabs}
\newcolumntype{C}{>{\centering\arraybackslash}X} % centered version of "X" type
\usepackage[obeyspaces]{url}
\usepackage{courier}
\DeclareMathOperator*{\argmin}{arg\,min}

\begin{document}

\title{Automatic Calibration of a Six-Degrees-of-Freedom Pose Estimation System}

\author{
\IEEEauthorblockN{Wouter Jansen, Dennis Laurijssen, Walter Daems, Jan Steckel}
\IEEEauthorblockA{University of Antwerp, Faculty of Applied Engineering - CoSys Lab\\
Groenenborgerlaan 171, Antwerp, Belgium\\
and the Flanders Make Strategic Research Centre, Lommel, Belgium\\
jan.steckel@uantwerpen.be}
}

\maketitle

\begin{abstract}
Systems for estimating the six-degrees-of-freedom human body pose have been improving for over two decades. Technologies such as motion capture cameras, advanced gaming peripherals and more recently both deep learning techniques and virtual reality systems have shown impressive results. However, most systems that provide high accuracy and high precision are expensive and not easy to operate. Recently, research has been carried out to estimate the human body pose using the HTC Vive virtual reality system. This system shows accurate results while keeping the cost under a 1000 USD. This system uses an optical approach. Two transmitter devices emit infrared pulses and laser planes are tracked by use of photo diodes on receiver hardware. A system using these transmitter devices combined with low-cost custom-made receiver hardware was developed previously but requires manual measurement of the position and orientation of the transmitter devices. These manual measurements can be time consuming, prone to error and not possible in particular setups. We propose an algorithm to automatically calibrate the poses of the transmitter devices in any chosen environment with custom receiver/calibration hardware. Results show that the calibration works in a variety of setups while being more accurate than what manual measurements would allow. Furthermore, the calibration movement and speed has no noticeable influence on the precision of the results.
\end{abstract}

\IEEEpeerreviewmaketitle

\section{Introduction}
\begin{figure}[!t]
\centering
\includegraphics[width=\linewidth]{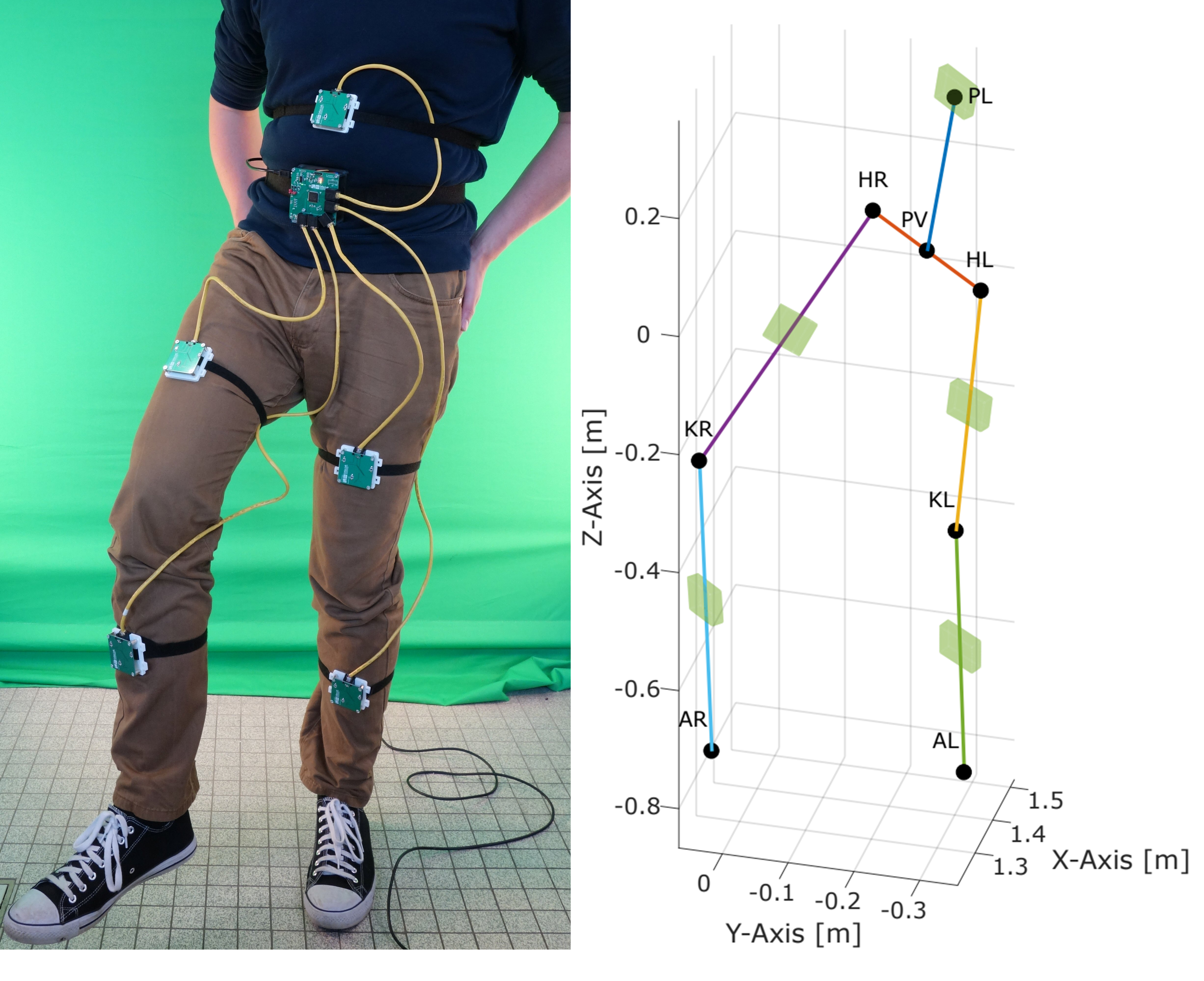}
\caption{Example of an experiment using the custom low-cost receiver hardware attached to the limbs of a test subject. It shows the representation of the human body pose on the right using tracking data obtained with the Lighthouses and the receiver hardware \cite{Laurijssen2017}.}
\label{fig:Laurijssen2017fig}
\end{figure}

In the last two decades, human body pose estimation has seen significant technological advances, predominantly due to the movie and video game industries, who have, for instance, been using specialised suits with passive (reflective coating) or active (LED) markers to track the movement of the actors with multiple cameras. Afterwards this data can be used to create realistic animations. Furthermore, the video game industry has used gaming peripherals to track the player's movements and gestures, to  translate these into the game world. Devices such as the Wii Mote, Microsoft Kinect and PlayStation Move controller were popular only a few years ago.\\
Recently, Virtual Reality (VR) has received widespread attention. Systems such as the Oculus Rift, PlayStation VR and HTC Vive have allowed the user's gestures to be accurately tracked and translated into the 3D virtual world. These systems estimate the pose of the head-mounted display (HMD) and controllers that the user is holding.\\
Moreover, another popular approach that has been explored in recent years is estimating the human body pose from 2D images taken by static cameras \cite{Chen3DMatching,Andriluka2010MonocularDetection, ChenArticulatedRelations,Tompson2014Real-TimeNetworks, Cao}. Like the VR systems it offers a low cost solution to the estimation problem. \\
Likewise, at industrial scale other techniques are used for pose estimation for digital manufacturing and assembly. Technologies such as wMPS (workshop Measurement and Positioning System)\cite{Xiong2012WorkspacePlanes} and iGPS (indoor Global
Positioning System)\cite{YANG2010NovelBeams} use laser transmitter devices with different rotation speeds and photoelectric sensors on the receiver. However, as they often focus on the 2D pose-estimation of industrial robots and seek accuracy in different criteria, these techniques aren't well suited for low cost six-degrees-of-freedom (6-DoF) pose estimation of humans. \\
In general, accurately estimating the 6-DoF pose of animate objects, robots, humans and animals with low latency and low computational intensity at a reasonable price is not only of interest to these mentioned applications but to others as well. \\
For example, in the medical field, doctors could use it to carefully analyse the posture of their patient. While accurate and precise systems designed for these markets exist, they are often expensive and difficult to set up and operate \cite{Vlasic2007,Mirek2007}. Systems such as the Microsoft Kinect have been used in research for an alternative lower cost solution \cite{Gabel2012,Ganganath2012} but could not provide accurate and precise results when compared to the more expensive systems \cite{Pfister2014}.\\
Another system that is currently being used in research for tracking and pose estimation purposes is the HTC Vive \cite{Kleinschmidt2017,Islam2016,Yang2017,Paijens2017,Zheng2014,Niehorster2017,Laurijssen2017,Egger2017}. It uses a separate transmitter device called a Lighthouse that is placed within the environment and can help in tracking a subject using its dual-axis lasers that sweep the area. The tracked subject can use these laser sweeps to calculate its own position and orientation. Moreover, the Lighthouses are not limited to only work with the HTC Vive HMD and controllers. Any tracker hardware that can receive and analyse the infrared laser sweeps of the Lighthouses can use them to estimate the subject's pose. Furthermore, two Lighthouses can work together and synchronise automatically in order to achieve better results. This system operates automatically and could, therefore, be an ideal low-cost system for human body pose estimation that can be set up in a small amount of time. \\
The Lighthouses were used in combination with custom-made receiver hardware in previous research by Laurijssen et al. \cite{Laurijssen2017}. It was found that this setup could prove to be a low-cost system for precisely and accurately estimating the 6-DoF human body pose when multiple such receivers are attached to the limbs of a subject. The system can be seen in Figure \ref{fig:Laurijssen2017fig}.\\
However, it is not straightforward to set up. The system is dependent on the tracking software knowing the poses of the Lighthouses in order to calculate the tracked subject's pose within the environment. Manually measuring the pose of the Lighthouses can be a tedious and time consuming task, and inaccurate due to human error. It may even not be feasible in particular environments and require to avoid complex configurations. 
In this paper we describe a process for automatically calibrating the full 6-DoF pose of multiple Lighthouses within the environment using a probabilistic sensor fusion algorithm. We designed this method to be fast, robust and require no additional operator input. This will allow for a simpler and faster deployment of the system in any environment and enable accurate 6-DoF pose estimation for humans or other subjects.

\section{Lighthouse System}
\label{sec:lighthouseSystem}
\begin{figure}[!t]
\centering
\hbox{\hspace{5.8em}
\includegraphics[width=1.9in]{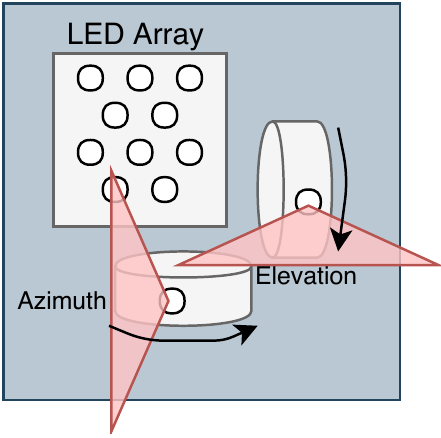}}
\caption{Illustration of the design of a Lighthouse transmitter from the front, showing the infrared sweeps in red.}
\label{fig:lighthouseSchematic}
\end{figure}

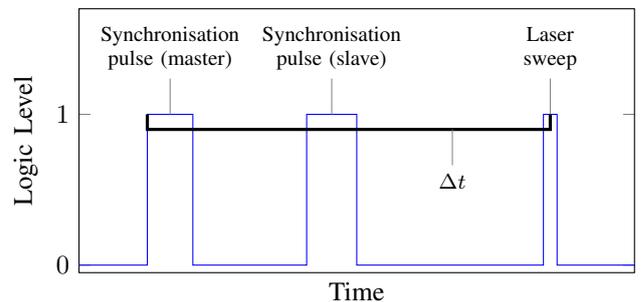
\begin{figure}[!t]
\begin{tikzpicture}
\begin{axis}[
    title style={align=center},
    title={Signal of the measurement cycle for\\a photo diode with two active Lighthouses},
    xlabel={Time},
    ylabel={Logic Level},
    height=2in,
    xmajorticks=false,
    ymin=-0.05,
    xmin=-1.2,
    xmax=11,
    ymax=1.7,
    ytick={0,1},
    grid style=dashed,
]
\addplot[color=blue] coordinates {(-10,0)(0.3,0)(0.3,1)(1.3,1)(1.3,0)(3.8,0)(3.8,1)(4.9,1)(4.9,0)(9,0)(9,1)(9.3,1)(9.3,0)(15,0)};
\addplot[very thick,color=black] coordinates {(0.3,1)(0.3,0.9)(9.15,0.9)(9.15,1)};
\node [coordinate,pin=above:{Synchronisation\\pulse (master)}] at (axis cs:0.8,1) {};
\node [coordinate,pin=above:{Synchronisation\\pulse (slave)}] at (axis cs:4.35,1) {};
\node [coordinate,pin=above:{Laser\\sweep}] at (axis cs:9.15,1) {};
\node [coordinate,pin=below:{$\Delta t$}] at (axis cs:7,0.9) {};
\end{axis}
\end{tikzpicture}
\caption{This figure shows an example of a signal for a photo diode on the receiver hardware. The time difference $\Delta t$ between the start of the synchronisation pulse and the middle of the laser sweep pulse ( the same principle for both horizontal and vertical sweeps) can be transformed to an angle given the rotation speed of the laser sweep. We will be referring to the two lighthouses as master and slave.}
\label{fig:photodiodesignal}
\end{figure}

The Lighthouse device has two rotating infrared lasers that sweep the environment alternately \cite{OliverKreylos}. Any tracker hardware with photo diodes that are sensitive to light of the right wavelength, can receive the emitted infrared light sweeps from these lasers.
The two infrared laser sweeps are perpendicular to one other as one rotates vertically and the other horizontally. The sweeps span \ang{120} each. Figure \ref{fig:lighthouseSchematic} shows the design of a Lighthouse. Between each laser sweep an infrared LED array within the Lighthouse flashes. This is used to optically synchronise two Lighthouses working simultaneously and more importantly to indicate the start of the next laser sweep. The time difference between the start of the synchronisation pulses and the middle of the laser pulse when they sweep across the photo diode can then be measured. An example of a measurement cycle is shown in Figure \ref{fig:photodiodesignal}. This cycle is repeated for both horizontal and vertical sweeps for both Lighthouse devices.\\ Given the constant rotational speed of the motors containing the mirrors that reflect the lasers, this measured time difference can be transformed to an angle. By performing this for both sweeps, the azimuth and elevation angles in reference to the origin of the Lighthouse can be obtained for each photo diode on the receiver hardware. The Lighthouses can operate up to a distance of \SI{7}{\m} or more depending on the conditions of the environment \cite{Niehorster2017}.\\

\section{Calibration Hardware Board}
\begin{figure}[!t]
\centering
\includegraphics[width=0.95\linewidth]{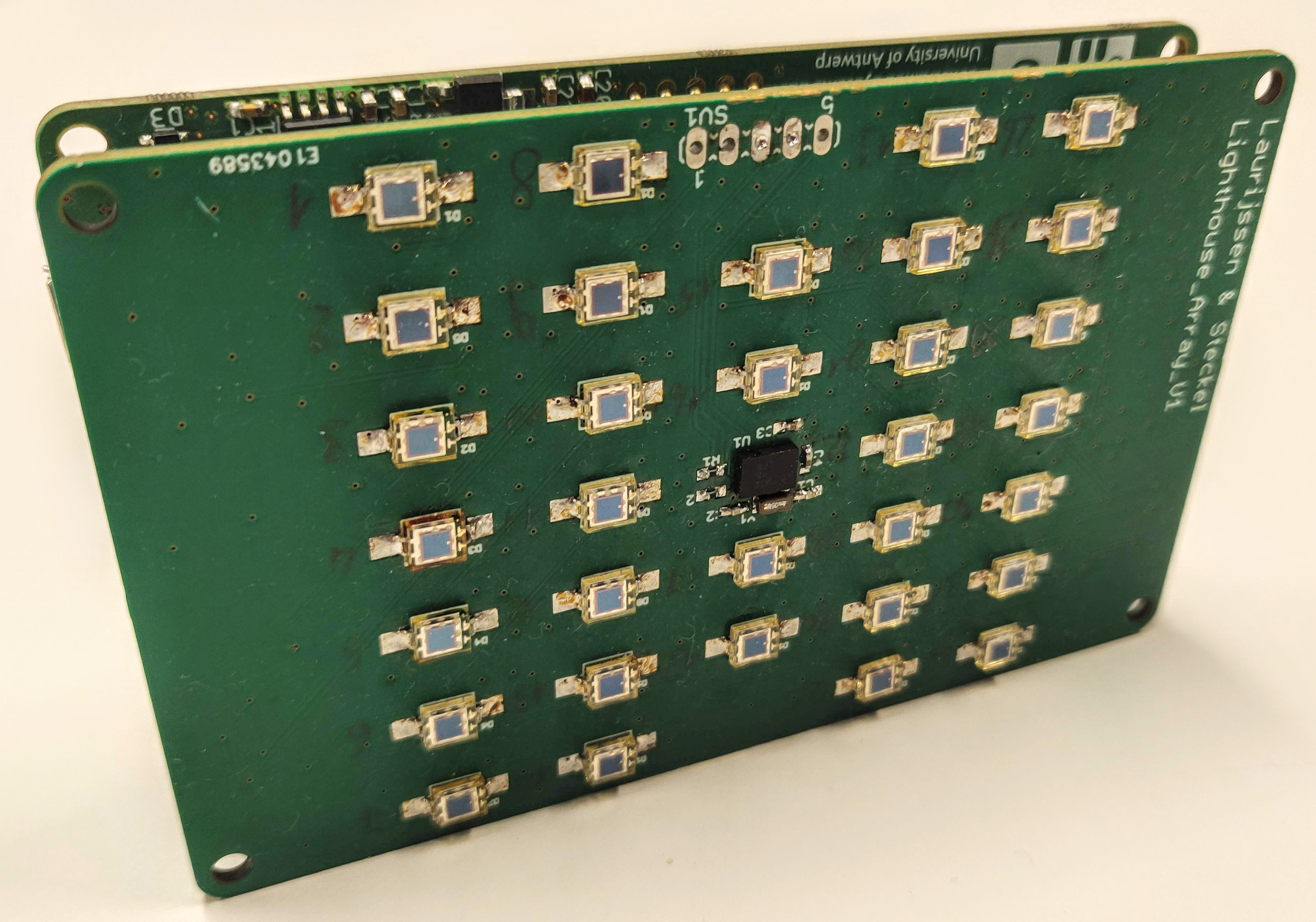}
\caption{The calibration hardware board showing the 32 photo diodes. This board is mounted on a second board containing the micro controller and its peripherals.}
\label{fig:calibrationboard}
\end{figure}

For performing the calibration we require specialised hardware. The previously developed receiver hardware by Laurijssen et al. was designed as modules to be attached to the limbs of the tracked subject \cite{Laurijssen2017}. Most of the hardware design can remain the same as a pose estimation measurement is comparable to a calibration measurement. However, to lower the ambiguity and to have data redundancy to account for lower quality measurements, a larger amount of sensor readings is required during the calibration. Furthermore, the amount of unknown variables is much larger during calibration than for pose estimation.\\
Therefore, the existing design was scaled up from 3 to 32 photo diodes (Vishay VBP104S), with accompanying trans-impedance amplifier, high-pass filter circuit and non-inverting op-amp for each photo diode. All photo diode GPIO connectors are continuously sampled at \SI{2}{\mega\hertz} by the microcontroller (STM32F429BET6). As there is no requirement for calibrating real-time, all data is stored in a on-board memory module (IS42S16320F-7TLI). When the measurement is finished the dataset can be transferred by USB using the FT232HL integrated circuit which uses the UART protocol for data transfer from the microcontroller. An initial implementation of this board can be seen in Figure \ref{fig:calibrationboard}. The calibration measurement is set to have a total duration of \SI{8}{\s}. Usage of this board showed that no jitter or other noise could be measured at the chosen sample rate, indicating that any measurement error solely results from quantization. 

\section{Measurement data reconstruction}
As stated before, the measurements are taken alternately due to the operation of the lasers. This means that a measurement will only contain either azimuth or elevation data. However to perform a proper pose calculation, both are required for a single measurement point. In order to make a measurement contain both angle types, a few options exist to reconstruct the measurement data. The two options we explored were interpolation and merging the interleaved measurements in pairs. When looking further at interpolation,  this as well can be done in different ways. A first approach was full interpolation where a measurement containing only azimuth angle data is complemented with linear interpolated elevation angle data based on the elevation measurements before and after that point in time. Similarly, when a measurement contains only elevation angle data, the azimuth data is found through interpolation. A second interpolation variant is to first calculate whether the measurements varied most in azimuth or elevation over the duration of the entire measurement. Based on this, the measurement type with the angle data that varied the most would have the other angle type interpolated. This means that the amount of measurement points would be halved in comparison to full interpolation. Nevertheless, this would be more accurate than the full interpolation as less interpolated data  would be used. While linear interpolation was used in both cases, a more accurate kinematic approach for taking into account the movement of the hardware board would be more optimal. However, this requires additional research of the movement during calibration which is out of the scope of this paper.
However, both interpolation and merging of the measurements are prone to inaccurate pose calculation when the calibration movement is done at a high speed. Each measurement cycle takes \SI{8.3}{\ms} \cite{OliverKreylos}. This starts with the synchronisation pulse as explained in Section \ref{sec:lighthouseSystem} and a short waiting period until the laser becomes visible at the start of the sweep at \SI{1.2}{\ms} into the cycle. The laser disappears again at the other end of the sweep at \SI{6.7}{\ms} into the cycle \cite{Yates}. This means that the maximum time difference between two consecutive measurements is \SI{13.8}{\ms}. Depending on the speed of the movement during calibration, the distance travelled in that maximum time frame can be of significance. If the movement of the receiver hardware is for example \SI{1}{\meter/\second}, it will result in a distance of \SI{1.3}{\cm} between measurements. If we then apply data reconstruction without prior knowledge of this speed, this could result in inaccurate pose calculations. Therefore, it is recommended that during calibration a low movement speed should be maintained in order to avoid inaccurate results. When comparing the interpolation and merge methods, we determined through experimentation with both simulation and real measurements that full interpolation proved to be the most accurate. However, the difference with the other methods was marginal.

\tdplotsetmaincoords{60}{120}
\begin{figure}[!t]
\centering
\begin{tikzpicture}[tdplot_main_coords]
  \draw[thick,->] (0,0,0) -- ( 2.5,0,0) node[anchor=south]{$x$};
  \draw[thick,->] (0,0,0) -- ( 0,2.5,0) node[anchor=west]{$y$};
  \draw[thick,->] (0,0,0) -- ( 0,0,2.5) node[anchor=north east]{$z$};  
  
  \draw[very thick,red] (0,0,0) -- (2,2,2) node[anchor=west]{P};
  \draw[very thick,blue] (0,0,0) -- (2,0,2) node[anchor=east]{P'};

  \draw[dashed,gray] (0,0,0) -- (2,2,0);
  \draw[dashed,gray] (2,2,0) -- (2,0,0);
  \draw[dashed,gray] (2,2,0) -- (0,2,0);
  \draw[dashed,gray] (2,2,2) -- (2,2,0);
  \draw[dashed,gray] (2,2,2) -- (2,0,2);
  \draw[dashed,gray] (0,0,2) -- (2,0,2);
  \draw[dashed,gray] (2,0,2) -- (2,0,0);
  \draw[dashed,gray] (0,2,2) -- (2,2,2);

  \tdplotdefinepoints(0,0,0)(0,0.5,0)(2,2,0);
  \tdplotdrawpolytopearc[thick,->]{1.5}{anchor=north}{$\theta$}
  \tdplotdefinepoints(0,0,0)(0,0,0.5)(2,0,2);
  \tdplotdrawpolytopearc[thick,->]{1.5}{anchor=east}{$\phi'$}
  \tdplotdefinepoints(0,0,0)(0,0,0.5)(2,2,2);
  \tdplotdrawpolytopearc[thick,->]{1.5}{anchor=west}{$\phi$}
\end{tikzpicture}
\caption{The elevation (\(\phi\)) and azimuth (\(\theta\)) angles as defined in a spherical coordinate system for a photo diode \(P\) are shown in this figure. However, when using the Lighthouse measurements, we are calculating the \(\phi'\) angle of the projected photo diode \(P'\) on the XZ-plane. Nevertheless, in this paper we will refer to this angle as the elevation angle for the sake of brevity.}
\label{fig:angleExplanationFig}
\end{figure}
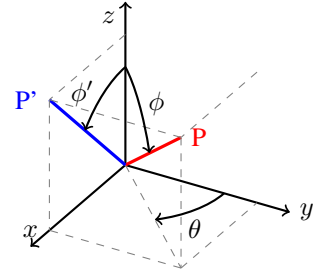

\section{Pose Estimation}
\subsection{Tracked receiver pose estimation}
\label{sec:trackedSubject}
Most academic research that uses the Lighthouse devices attempts to develop solutions to estimate the pose of the tracked subject and already has a known pose for the Lighthouse devices \cite{Kleinschmidt2017,Islam2016,Yang2017,Paijens2017,Zheng2014,Niehorster2017,Laurijssen2017,Egger2017}, when in fact we are trying to find both during calibration. However, the methods used to do this pose estimation are also required for the calibration. \\Therefore, we will focus finding the right approach in our context first. 
Firstly, it is important to state that while throughout this paper we speak of the elevation angle, in reality this is ambiguous. As Figure \ref{fig:angleExplanationFig} shows, it is the elevation angle of the photo diode projected on the XZ-plane that is measured. However we will continue to reference this angle as the elevation angle for brevity's sake. \\
When the dataset of a measurement with one or multiple Lighthouses and receiver hardware has been acquired, several methods have been suggested in academic literature for solving the pose of the tracked subject with that data. A first interesting approach is by direct geometric calculation as seen in research by Islam et al. (2016)\cite{Islam2016} and Kleinschmidt et al. (2017)\cite{Kleinschmidt2017}. A root-finding algorithm (Newton's method) is used in order to approximate the distance between the tracked subject and the Lighthouse. This method will minimise the distance error. The formulae to define this distance error come from solving a set of non-linear equations starting from trigonometric identities in the coordinate space of the Lighthouse. This estimated distance is combined with the measured azimuth and elevation angles for a direct calculation of the pose. However, in these papers it is assumed that both lasers are in the centre of the Lighthouse, which is not correct \cite{Alan2016}.\\
Next to the geometric approach, another possibility is using Perspective-n-Point (PnP), a computer vision method for solving poses. The PnP method sees the Lighthouse as a camera, and the tracked subject is seen as a projection on the camera's lens. Similarly as the geometric approach, the angles measured can be used for measuring the direction of the tracked subject on the projection. Minimisation techniques are used in order to solve the distance, similarly as has to be done for the geometric approach. Algorithms such as the Gauss-Newton's algorithm are ideal for solving the distance. The PnP algorithm is usually implemented with either an iterative or non-iterative approach or a combination of both to improve the result \cite{Lepetit2009EPnP:Problem}. Pose estimation using PnP with Lighthouse devices has been successfully applied by Zheng et al. (2014) \cite{Zheng2014} and Yang et al. (2017)\cite{Yang2017}. Similarly as with the geometric approach, the lasers are in the centre of the camera's perspective. It remains to be shown how the accuracy of the pose calculation is influenced by this assumption.\\
An alternative approach was used in our solution. First, we developed a simulation algorithm that can generate the dataset containing the azimuth ($\theta$) and elevation ($\phi$) angles for each photo diode on the calibration hardware board from the perspective of a Lighthouse. This algorithm requires as input the Lighthouse 6-DoF pose $\vec{L}$ defined by a position ($x, y, z$) and a rotation ($\alpha, \beta, \gamma$) (and subsequently the position of the lasers within the Lighthouse) and the position of each of the 32 photo diodes on the hardware calibration board. These 32 points will define a pose $\vec{P}$ of the hardware calibration board. This pose can be calculated using the positions of the photo diodes as their array configuration is known a priori. For calculating the angles, as seen in Equation \ref{eq:angleCalc}, the four-quadrant inverse tangent (\texttt{atan2}) is used where the corresponding laser is considered to be the origin of the system for each calculation.

\begin{equation}
\label{eq:angleCalc}
\begin{aligned}
\theta_i = \texttt{atan2}(\vec{P_{y}}, \vec{P_{x}})\\
\phi_i = \texttt{atan2}(\vec{P_{z}}, \vec{P_{x}})
\end{aligned}
\end{equation}
With these equations the same dataset can be generated as a real measurement would with the calibration hardware board and a Lighthouse. This yields a matrix of azimuth and elevation data $\mathbf{C}(\vec{L},\vec{P})$:

\begin{equation}
\label{eq:functionDefEq}
\begin{aligned}
\mathbf{C}(\vec{L},\vec{P}) =\begin{bmatrix}
    \theta_1      & \phi_1  \\
    \theta_2      & \phi_2 \\
    \vdots      & \vdots \\
    \theta_{32}   & \phi_{32}
\end{bmatrix}
\end{aligned}
\end{equation}
When the same data is obtained from a real measurement using the receiver hardware, we will refer to it as $\mathbf{M_t}$.
The pose estimation problem is nonlinear and can be solved using optimisation. We use the Nelder-Mead simplex optimisation algorithm which we can write as:

\begin{equation}
\label{eq:minimizationEq}
\begin{aligned}
{\vec P_*},{\epsilon_P}  = \argmin_{\vec P_i} K_1(\vec P_{0},\vec L,\mathbf{M_t})
\end{aligned}
\end{equation}

\begin{equation}
\label{eq:lossFunctionEq}
\begin{aligned}
K_1 = \begin{Vmatrix} \mathbf{C}(\vec L,\vec P_e) - \mathbf{M_t} \end{Vmatrix}_F^2
\end{aligned}
\end{equation}

Where ${\vec P_*}$ is the solution of the minimisation problem, the estimated pose of the tracked receiver and $\epsilon_P$ the remaining estimation error. $\vec P_{0}$ is the initial guess to start with. This initial pose is calculated by taking the mean of the elevation and azimuth angles of $\mathbf{M_t}$. The loss function $K_1$ to use in the optimisation is shown in Equation \ref{eq:lossFunctionEq}. With $\begin{Vmatrix}\cdot\end{Vmatrix}_F$ denoting the Frobenius norm. $\vec P_e$ is the approximated pose of the receiver hardware which we are trying to estimate. 

\subsection{Lighthouse pose estimation}
\label{sec:ligthouseEstimation}
A similar method is used in order to find the reverse, the 6-DoF pose of a Lighthouse device in the environment. A calibration with the hardware calibration board with $N$ measurement points (after measurement data reconstruction) will result in $N$ datasets $\mathbf{M_{t,n}}$, where $n$ is the measurement index. Together with the 6-DoF pose of the hardware calibration board for each measurement $\vec P_{n}$, this can once more be solved as a nonlinear optimisation problem, but now for the pose estimation of the Lighthouse device $\vec L$. 

\begin{equation}
\label{eq:minimization2Eq}
\begin{aligned}
 {\vec L_*},{\epsilon_L}  = \argmin_{\vec L} K_2(\vec P_{n},\vec L_0,\mathbf{M_{t,n}})
\end{aligned}
\end{equation}

\begin{equation}
\label{eq:lossFunction2Eq}
\begin{aligned}
K_2 = \sum_{n=1}^{N} \begin{Vmatrix} \mathbf{C}(\vec L_e,\vec P_{n}) - \mathbf{M_{t,n}} \end{Vmatrix}_F^2
\end{aligned}
\end{equation}

The minimisation function is altered in order to use an initial value of the Lighthouse pose $\vec L_0$. The result is the estimated Lighthouse 6-DoF $\vec L_*$. How the initial value is calculated is explained in the next section when the calibration algorithm is discussed. A new objective function $K_2$ is defined as seen in Equation \ref{eq:lossFunction2Eq}, summing over all $N$ measurements.

\section{Proposed Automatic Calibration Algorithm}
The overall goal of the proposed algorithm is to find the positional and rotational difference between two Lighthouses. The different steps of the complete calibration algorithm are explained in sequence. 
\subsection{Pose estimation measurements for master Lighthouse}
\label{subsec:masterEst}
First, for every measurement made from the first Lighthouse (master) resulting in a dataset $\mathbf{M_t}$ as seen in Equation \ref{eq:functionDefEq}, a pose estimation is performed as discussed in Section \ref{sec:trackedSubject} using the minimisation described in Equation \ref{eq:minimizationEq}. The Lighthouse pose $\vec L$ is set at the origin of the system. Executing this for every measurement point will result in a pose estimation of the entire measurement path of the calibration board in reference to the the master Lighthouse at the origin of the coordinate system. This step is shown in Figure \ref{fig:algorithmOne}.
\subsection{Pose estimation measurements for slave Lighthouse}
\label{subsec:slaveEst}
\begin{figure}[!t]
\centering
\includegraphics[width=\linewidth]{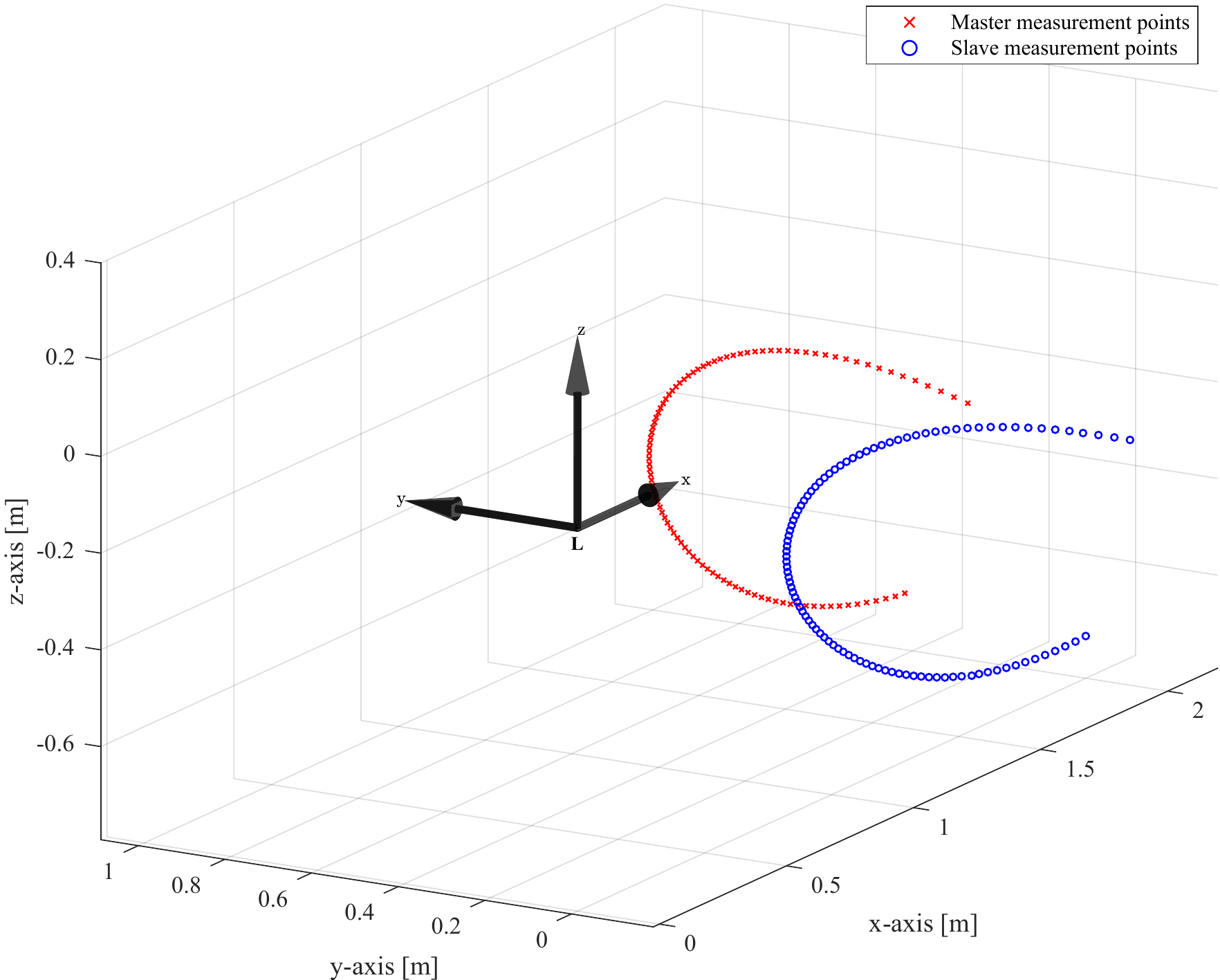}
\caption{P estimation for each measurement point of both Lighthouses. This simulated example shows a half circle shape. Both Lighthouses are placed at the origin of the system.}
\label{fig:algorithmOne}
\end{figure}

The same procedure is followed for the second Lighthouse (slave) as shown in Figure \ref{fig:algorithmOne}. Once more, the Lighthouse pose $\vec L$ is placed at the origin of the system resulting in a pose estimation of the entire measurement path of the calibration board in reference to the slave Lighthouse positioned in the origin of the system.

\subsection{Measurement interpolation}
While the two estimated measurement paths will be very similar as can be seen in figure \ref{fig:algorithmOne}, they are not the same. There is a clear visible offset in both the position and rotation of both paths. Another issue is that the calibration measurements of the two lighthouses don't occur at the same time. They are performed alternately as discussed in Section \ref{sec:lighthouseSystem}. Consequently, they have a limited correlation between them if we would compare the individual points.\\ However, knowing the alternated measurement order they occurred, it is possible to interpolate the measurement path to contain the measurement poses of the other lighthouse. In this way, both measurement paths will have an equal amount of points corresponding to the same moments in time, which can be used to find a correlation between the two measurement paths. We used linear interpolation to achieve our results. 

\begin{figure}[!t]
\centering
\includegraphics[width=\linewidth]{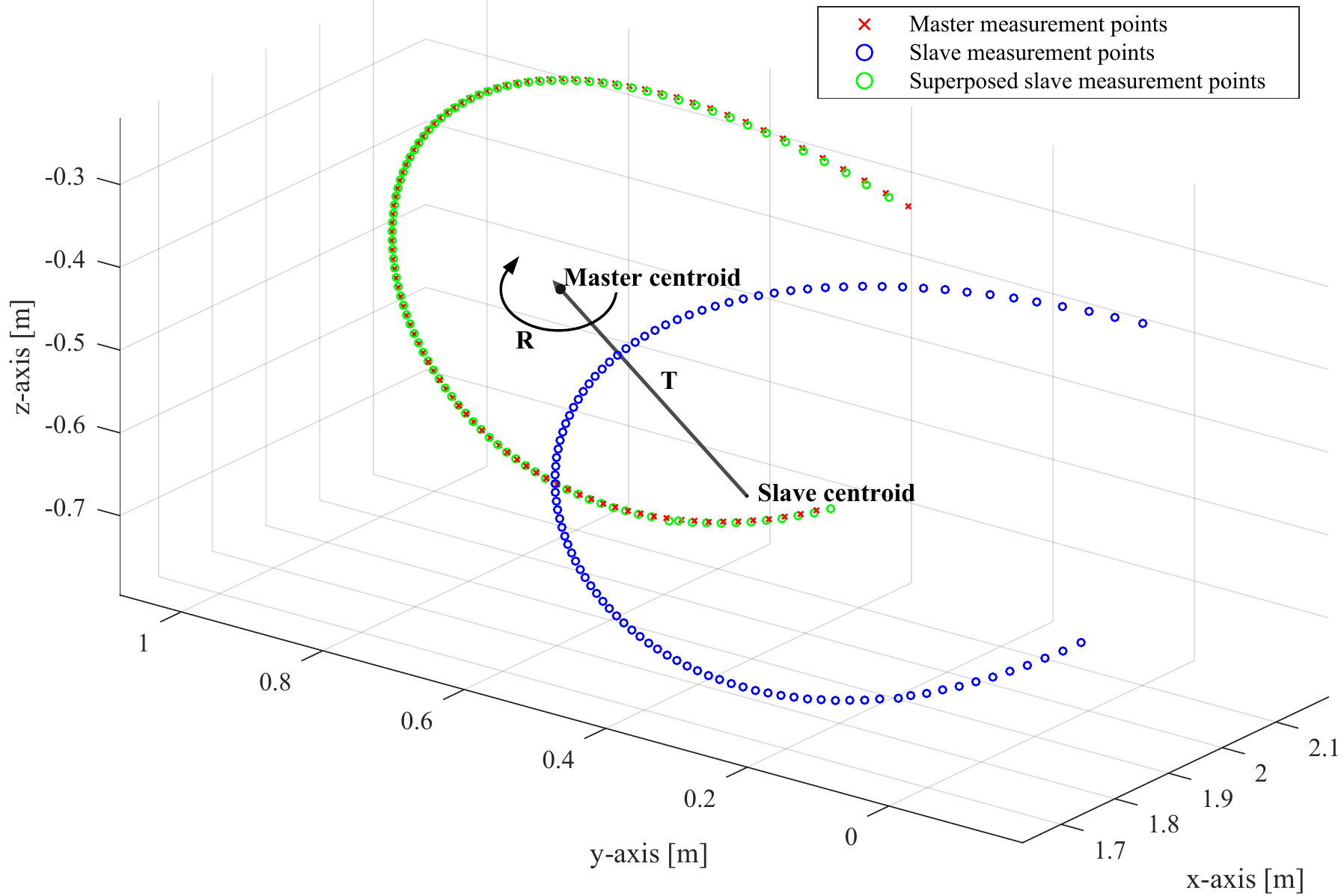}
\caption{The superposition of the measurement path of the slave Lighthouse to the master Lighthouse's measurement path using the Kabsch algorithm. The arrows illustrate the translation vector $\mathbf{T}$ and rotation matrix $\mathbf{R}$ being applied.}
\label{fig:algorithmTwo}
\end{figure}

\subsection{Superposing the measurements}
While the two estimated measurements paths are now similar in time, there are still not in position and rotation. Therefore, we use superposition using the Kabsch algorithm \cite{Kabsch1976} to solve this. First, the Kabsch algorithm will translate the measurement path of the slave Lighthouse so that its centroid will coincide with that of the measurement path of the master Lighthouse. This will solve the positional offset mentioned previously. Afterwards, the measurement path of the slave lighthouse is rotated to have the same orientation as that of the master Lighthouse. In order to achieve this, the Kabsch algorithm calculates the optimal rotation matrix using a singular value decomposition routine. The principle is illustrated in Figure \ref{fig:algorithmTwo}.

\begin{figure}[!t]
\centering
\includegraphics[width=\linewidth]{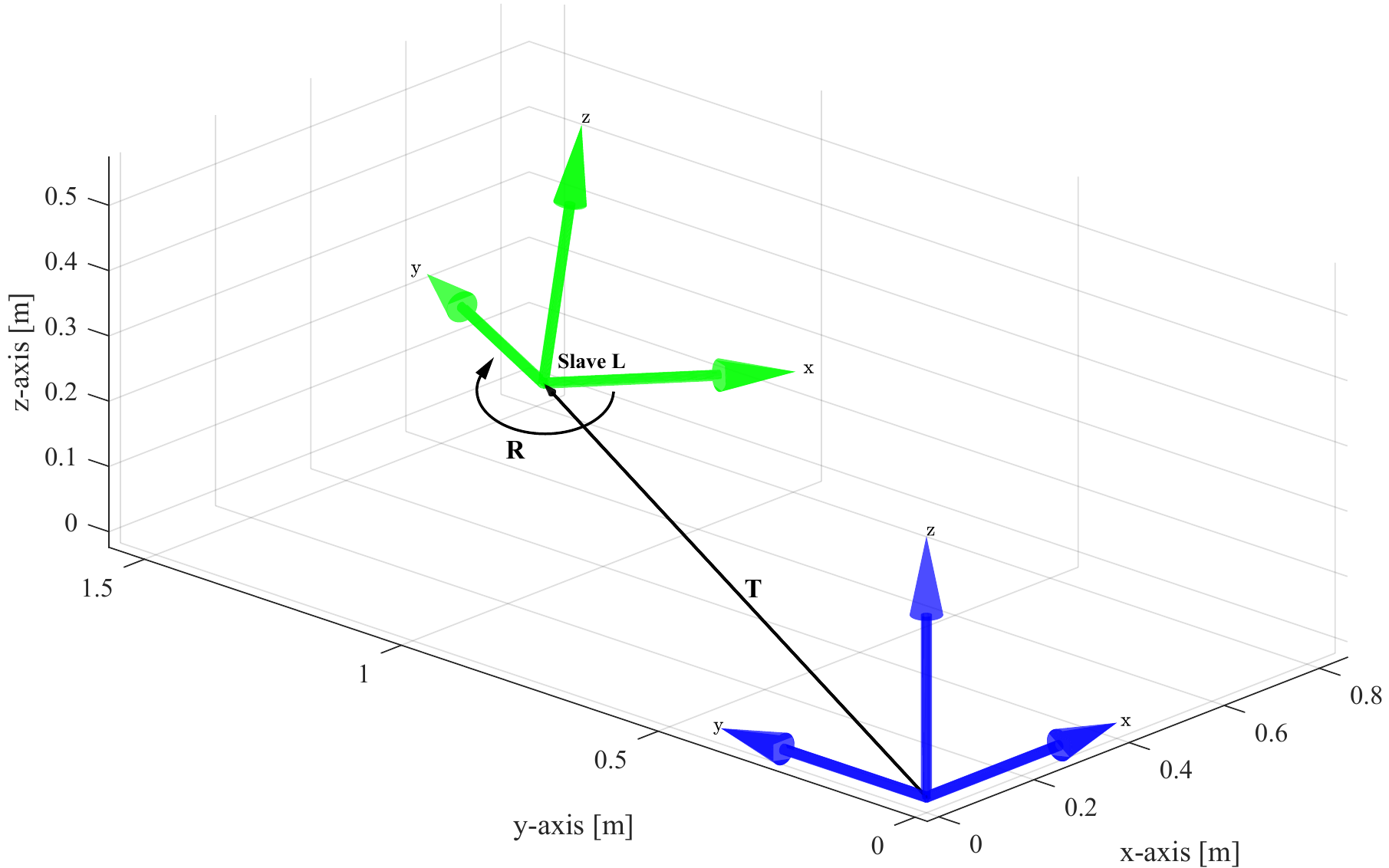}
\caption{The translation vector $\mathbf{T}$ and rotation matrix $\mathbf{R}$ calculated through the Kabsch algorithm being applied and giving an estimation of the slave Lighthouse pose. This estimated pose can be further improved with an additional pose estimation.}
\label{fig:algorithmThree}
\end{figure}

\subsection{Initial Pose Estimation of the second Lighthouse}
When applying the same calculated rotation matrix $\mathbf{R}$ and translation vector $\mathbf{T}$ to the slave Lighthouse pose (which is at the origin of the system), an initial estimated pose for the slave Lighthouse is found. This step is shown in Figure \ref{fig:algorithmThree}. When executing the initial pose estimation of the measurements as described in Subsections \ref{subsec:masterEst} and \ref{subsec:slaveEst}, the minimisation function used in this procedure, which is shown in Equation \ref{eq:minimizationEq}, will return the remaining error $\epsilon_P$ for each measurement point. These error values will be used as weights for the Kabsch algorithm to put more emphasis on measurements points that were accurately estimated and less emphasis on those with a higher remaining error.

\subsection{Final Pose Estimation of the second Lighthouse}
Lastly, the estimation method for a Lighthouse detailed in Section \ref{sec:ligthouseEstimation} is applied. Equation \ref{eq:minimization2Eq} is used to estimate the pose of the slave Lighthouse. The estimated pose which was calculated in the previous step is used as initial value $\vec L_0$ for the minimisation here. This is the final pose estimation of the slave Lighthouse in reference to the master Lighthouse. Depending on the accuracy of the measurement data, this result will differ very little from the initial estimation from the previous step.

\section{Results}
\begin{figure}[!t]
\centering
\includegraphics[width=\linewidth]{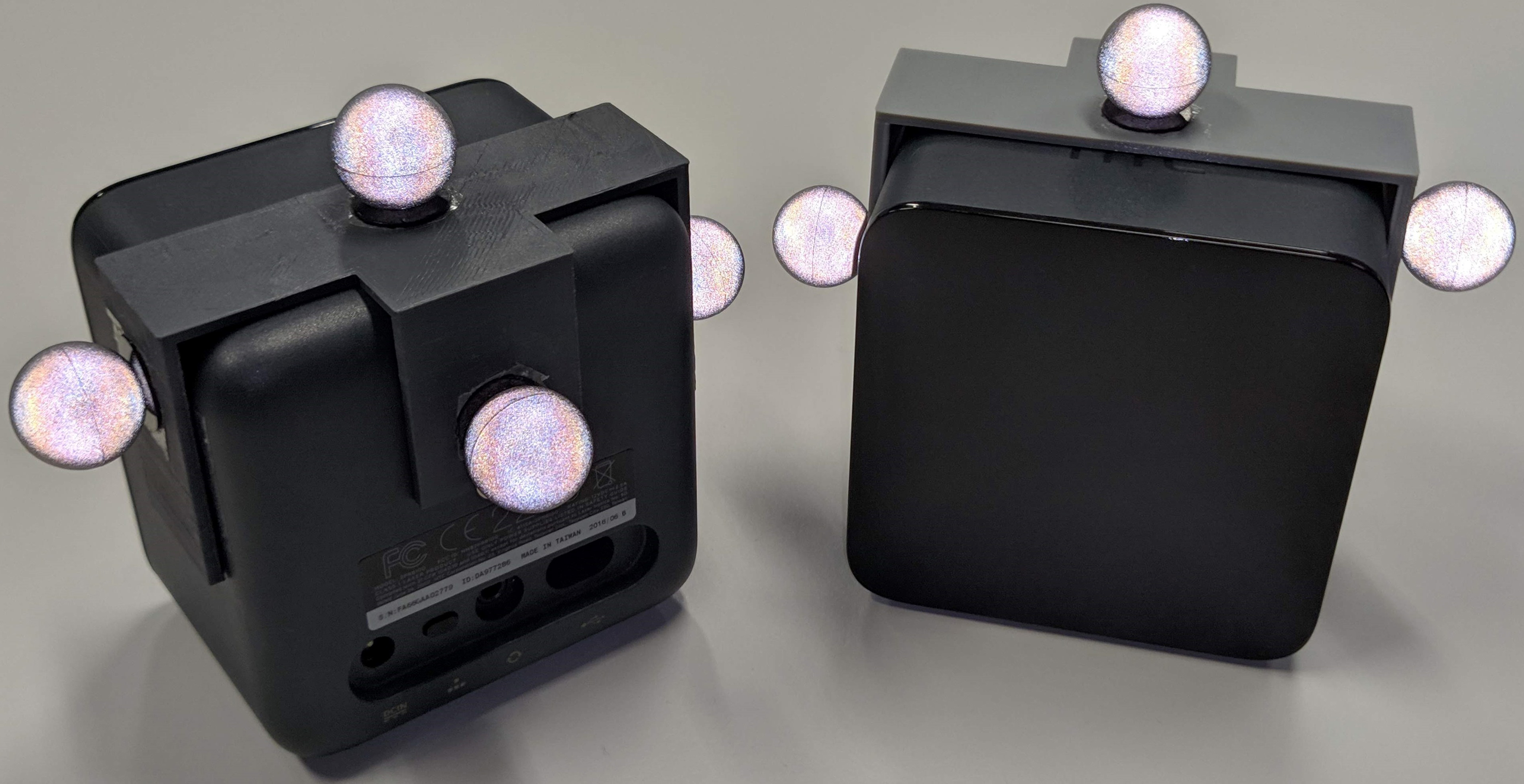}
\caption{Two Lighthouse devices with Qualisys markers used during motion capture for validating the results. A 3D-printed mount was used to attach the markers in order to be able to accurately track the Lighthouse from a defined origin point in reference to the mount.}
\label{fig:lighthouseMarkers}
\end{figure}

To validate the algorithm, eight different and varied setups using two Lighthouse devices were created, each setup being different in terms of rotation and distance between the two Lighthouses. Five measurements were performed for each setup. Each measurement was varied in motion pattern, distance to the Lighthouses and movement speed. They were chosen at random and in no particular order. During early experiments we were unable to find a correlation between the setup of the calibration (the motion pattern, variation in motion speed and the overall setup of the devices) and the end result. This allowed us to choose random start parameters for the results shown in this paper. The distances between the Lighthouse devices varied between \SI{1}{\m} to \SI{4}{\m}. Distances between the calibration hardware to the lighthouse devices varied between \SI{2}{\m} to \SI{6}{\m}. \\
The positional and rotational difference between the two Lighthouses is estimated using our proposed algorithm and compared to measurements of a Qualisys motion capture system\cite{Qualisys}. This system uses passive markers with multiple high-speed digital cameras working in the infrared spectrum to track the markers. In Figure \ref{fig:lighthouseMarkers} two Lighthouse devices can be seen with the Qualisys markers attached using a 3D-printed mount. Using the Qualisys system provides a more accurate validation compared to manual measurement for validation. The validation with the Qualisys system is accurate up to less than \SI{1}{mm}. 

\begin{table}[hbt!]
 \caption{Mean Absolute Error Results}
\label{table:resultsTableMAE}
\centering
\begin{tabular}{c|c|c|c|c|c|c}
\toprule
Setup   & X (mm)    & Y (mm)    & Z (mm)    & $\alpha$ ($^\circ$)   & $\beta$ ($^\circ$)    & $\gamma$ ($^\circ$)\\ 
\midrule
1       &  9.9      & 23.0      & 32.9      & 1.35                  & 3.09                  & 0.80 \\  
2       &  4.5      & 27.4      &  7.2      & 0.20                  & 1.31                  & 0.96 \\ 
3       &  7.5      & 15.4      & 58.1      & 0.57                  & 3.18                  & 2.35 \\ 
4       &  6.7      &  8.4      & 19.1      & 0.40                  & 3.08                  & 1.88 \\ 
5       & 22.6      & 25.4      & 47.4      & 3.53                  & 1.70                  & 3.31 \\ 
6       & 30.4      & 41.5      & 39.5      & 0.42                  & 1.32                  & 1.24 \\ 
7       & 27.5      &  8.6      & 12.1      & 0.35                  & 0.45                  & 0.24 \\ 
8       & 55.4      &  8.3      & 18.2      & 2.72                  & 0.20                  & 1.10 \\ 
\bottomrule
Overall & 20.6      & 19.7      & 29.3      & 1.19                  & 1.79                  & 1.43 \\ 
\end{tabular}
\end{table}

\begin{table}[hbt!]
 \caption{Standard Deviation Results}
\label{table:resultsTableSD}
\centering
\begin{tabular}{c|c|c|c|c|c|c}
\toprule
Setup   & X (mm)    & Y (mm)    & Z (mm)    & $\alpha$ ($^\circ$)   & $\beta$ ($^\circ$)    & $\gamma$ ($^\circ$)\\ 
\midrule
1       &  8.9      & 25.0      & 32.8      & 0.45                  & 1.09                  & 0.89 \\  
2       &  3.5      & 29.0      &  6.1      & 0.22                  & 0.19                  & 0.82 \\ 
3       &  9.5      &  7.2      &  9.2      & 0.39                  & 0.20                  & 0.55 \\ 
4       &  4.1      &  6.7      & 23.4      & 0.43                  & 0.71                  & 0.21 \\ 
5       & 16.9      &  8.1      &  6.7      & 0.25                  & 0.40                  & 0.47 \\ 
6       & 10.0      & 13.1      & 14.6      & 0.20                  & 0.44                  & 0.52 \\ 
7       &  1.4      & 10.1      & 15.6      & 0.40                  & 0.55                  & 0.28 \\ 
8       & 16.9      &  6.6      &  6.7      & 0.23                  & 0.23                  & 0.45 \\ 
\bottomrule
Average & 8.90      & 13.2      & 14.4      & 0.32                  & 0.47                  & 0.52 \\ 
\end{tabular}
\end{table}

\begin{table}[hbt!]
 \caption{Simulation Results}
\label{table:resultsSimulation}
\centering
\begin{tabular}{c|c|c|c|c|c|c}
\toprule
            & X (mm)    & Y (mm)    & Z (mm)    & $\alpha$ ($^\circ$)    & $\beta$ ($^\circ$)     & $\gamma$ ($^\circ$)\\ 
\midrule
MAE         &  6.9      &  9.3      &  8.5      & 0.10                  & 0.17                  & 0.16 \\ 
Average SD  &  8.9      & 13.5      &  9.3      & 0.11                  & 0.20                  & 0.22 \\ 
\end{tabular}
\end{table}

The accuracy of the system can be assessed by looking at the Mean Absolute Error (MAE) between the motion tracked result with Qualisys and the estimated result with our proposed algorithm. These results are shown in Table \ref{table:resultsTableMAE} for every setup. The standard deviation allows assessing the precision of the algorithm, which can be found in Table \ref{table:resultsTableSD}. Furthermore, the algorithm was tested in simulation, where a realistic sample error was simulated into the data. Simulated measurement data was created for ten setups of various Lighthouse poses. Five measurements were performed for each setup. Moreover, each measurement was varied in motion pattern, distance to the Lighthouses and movement speed  as it was done for the real measurements. The results of these simulations can be found in Table \ref{table:resultsSimulation}. With both the overall MAE and average average standard deviation being given. A discrepancy between reality and simulation in the results can be observed. While the precision is similar, the accuracy in simulation is better. Admittedly, a physical and mathematical difference between the real and simulated performance is still present. Sample error being incorrectly simulated does not account for this difference in accuracy alone. The simulation was helpful for testing various iterations of the algorithms but obviously does not accurately represent the accuracy of the real system. 

\section{Conclusion \& Future work}
The results show that an automatic calibration provides decent results in every tested configuration. Furthermore, the precision over multiple calibrations of the same configuration shows that movement and speed of the measurement has no noticeable impact on the results. Additionally, not every configuration allowed for an accurate manual measurement due to the distance and rotational difference between the Lighthouses. \\
Moreover, the entire calibration does not last longer than a few minutes on a laptop PC with an Intel Core i5-8250U CPU. This means that the calibration can be easily repeated when changes occur to the configuration of the Lighthouse devices. An important experiment that now can be performed with this calibration method is to look for the impact of the accuracy and precision of the calibration on the pose estimation of the tracked subject afterwards. \\
Furthermore, a comparison could be made between our system using Lighthouse devices and the systems using static cameras with resulting 2D images to derive full 3D human body poses as discussed in the introduction. Those second type of systems need to be calibrated as well by using computer vision techniques such as the PNP method as described in \ref{sec:trackedSubject}. While this is out of the scope of this paper, comparing the systems in the accuracy of the resulting human pose estimations, cost of the system and required computing power could validate the usage of these systems between one another and compared to more expensive commercial available solutions. \\
One of the remaining issues that can be further explored with the current system for automatic calibration is that a direct correlation between the data of the two lighthouses is absent. As discussed in previous sections, interpolated measurement points and superposition techniques are being used to solve this problem. We believe that additional sensor data could prove to be more successful. Therefore, we suggest integrating a 9-DoF Inertial Measurement Unit (IMU) in the hardware. This addition would help to further alleviate the problem of correlating measurement points between both Lighthouses. Applying sensor fusion and filters between IMU and Lighthouse measurements is how Valve and HTC achieve the pose estimation \cite{YatesAlen, Yates}.\\
Moreover, we suggest to further improve the accuracy of the system by researching additional properties of the Lighthouse devices. More notably the intrinsic parameters of the lasers. Parameters such as their fabrication offsets in curvature of the sweep plane, tilting of the laser beam and their position within the Lighthouse device. \\
Providing further testing and research, by more accurately estimating these parameters of the lasers during the calibration process can be expected to further improve accuracy of the calibration.\\

\bibliographystyle{IEEEtran}
\bibliography{bib.bib}

\end{document}